\documentclass[letterpaper, 10 pt, conference]{ieeeconf}  

\IEEEoverridecommandlockouts                              

\usepackage{siunitx}
\usepackage{layouts}
\usepackage{scalerel}
\usepackage{graphicx}
\usepackage{standalone}
\usepackage{pgfplots}

\let\labelindent\relax
\usepackage{enumitem}
\usepgfplotslibrary{groupplots, statistics}
\usepackage{tikz}
\usetikzlibrary{shapes,backgrounds,shapes.geometric}
\usepackage{circuitikz}
\tikzset{font={\fontsize{9pt}{9}\selectfont}}
\usepackage{wrapfig}
\usepackage{multirow}
\usepackage{makecell}
\usepackage{amsmath,amssymb,amsfonts}
\usepackage{bm}
\usepackage{subfigure}

\usepackage{jabbrv}

\usepackage{tabularx}
\usepackage{makecell}
\usepackage{multirow}
\usepackage{booktabs}
\newcolumntype{Z}{>{\centering\arraybackslash}X}
\makeatletter
\let\NAT@parse\undefined
\makeatother
\usepackage{hyperref}  

\title{\LARGE \bf
TrajFlow: Learning Distributions over Trajectories \\ for Human Behavior Prediction
}

\author{Anna M\'esz\'aros$^*$, Julian F. Schumann, Javier Alonso-Mora, Arkady Zgonnikov, and Jens Kober
\thanks{*Corresponding author}
\thanks{This research was supported by NWO-NWA project “Acting under uncertainty” (ACT), NWA.1292.19.298.}
\thanks{All authors are with the Cognitive Robotics Department,
        TU Delft, 2628 CB Delft, The Netherlands
        {\tt\small \{A.Meszaros, J.F.Schumann, J.AlonsoMora, A.Zgonnikov, J.Kober\}@tudelft.nl}}%
}

\begin{document}

\maketitle
\thispagestyle{empty}
\pagestyle{empty}

\begin{abstract}

Predicting the future behavior of human road users is an important aspect for the development of risk-aware autonomous vehicles. 
While many models have been developed towards this end, effectively capturing and predicting the variability inherent to human behavior still remains an open challenge. 
This paper proposes TrajFlow---a new approach for probabilistic trajectory prediction based on Normalizing Flows.
We reformulate the problem of capturing distributions over trajectories into capturing distributions over abstracted trajectory features using an autoencoder, simplifying the learning task of the Normalizing Flows. 
TrajFlow outperforms state-of-the-art behavior prediction models in capturing full trajectory distributions in two synthetic benchmarks with known true distributions, and is competitive on the naturalistic datasets ETH/UCY, rounD, and nuScenes.
Our results demonstrate the effectiveness of TrajFlow in probabilistic prediction of human behavior.

\end{abstract}

\section{INTRODUCTION}
Autonomous vehicles (AVs) have become an important field of research due to many promised benefits which include, but are not limited to, improved safety, accessibility, as well as reduced traffic congestion~\cite{brar2017impact, meyer2017autonomous, pisarov2021future}.
Yet they are still not widespread, in big part due to their inability to effectively resolve interactions with humans~\cite{milford2019self, brown2023halting}.
Being able to reliably and accurately predict human behavior would allow for more efficient and safe AV path planning~\cite{schumann2023benchmarking}.

\begin{figure}
    \centering
    \includegraphics[width=0.8\linewidth, trim={3.5cm 0 3.5cm 1.5cm}, clip]{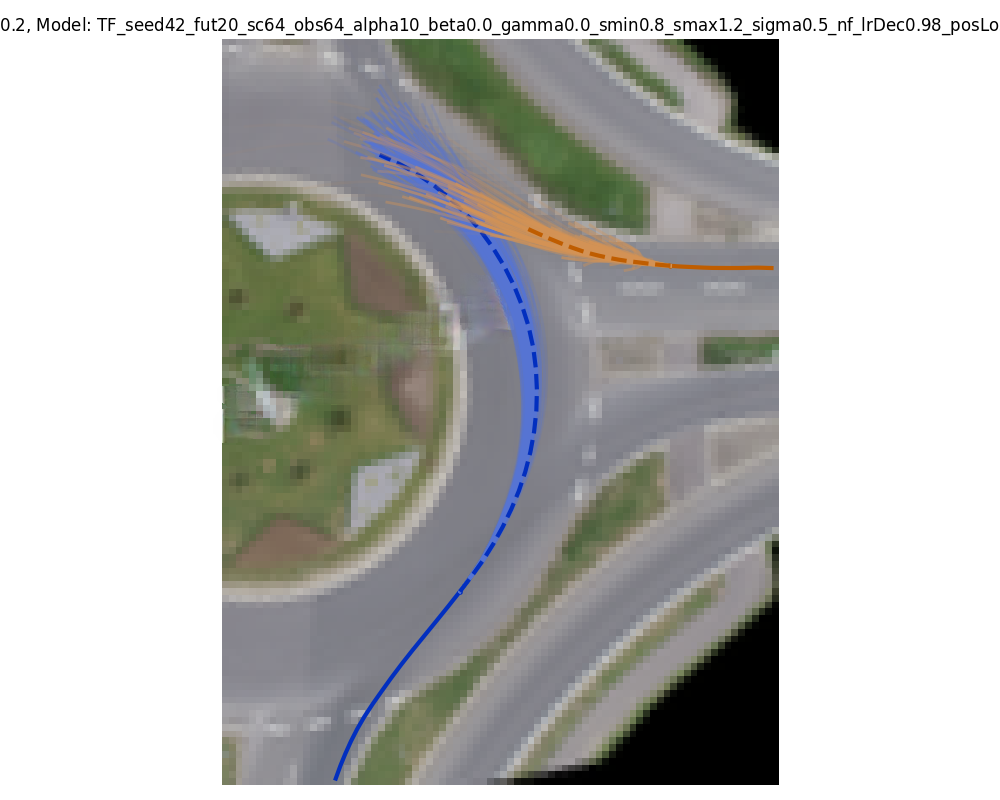}
    \vspace{-2mm}
    \caption{An example of the predictions generated by TrajFlow on the rounD dataset. Level of opacity indicates the likelihood of a given prediction.}
    \vspace{-4mm}
    \label{fig:frontPage}
\end{figure}

However, predicting human behavior in traffic is complicated by the fact that such behavior is generally not deterministic, but instead stochastic, with potentially complex and multi-modal distributions~\cite{ferro2020stochastic}. 
An example of such multi-modality can be seen at roundabouts, 
where vehicles have the option to enter the roundabout directly or to wait for an oncoming car to pass.
While these two options are the most obvious high-level behaviors, there can also be other distinct modes, such as deciding whether or not to slow down before entering the roundabout (Fig.~\ref{fig:frontPage}). Such modes are scenario-dependent and may get overlooked by methods that rely on a predefined number of modes~\cite{varadarajan2022multipath++, messaoud2021trajectory}.

Several methodologies for providing probabilistic predictions over traffic agents' future trajectories have been proposed, ranging from Gaussian Mixture Models (GMMs)~\cite{varadarajan2022multipath++, messaoud2021trajectory} to generative networks.
Generative networks such as Generative Adversarial Networks (GANs)~\cite{gupta2018social, amirian2019social}, Variational Autoencoder (VAE) based networks~\cite{salzmann2020trajectron++, yuan2021agentformer, mangalam2020not}, and diffusion models~\cite{gu2022stochastic}, are particularly interesting due to their potential to learn complex multi-modal distributions without specifying the number of expected modes, unlike methods that rely on GMMs.
While these state-of-the-art approaches already achieve good results in prediction accuracy, they have the fundamental problem of being trained to reproduce the \textit{only} true future trajectory available for each past trajectory in the dataset, thereby ignoring the underlying stochasticity of human behavior.
This training approach can result in mode collapse, which is especially problematic for GAN-based methods~\cite{gupta2018social, amirian2019social}.
Additionally, many state-of-the-art models predict distributions at individual time steps~\cite{salzmann2020trajectron++, de2021social}, ignoring the correlation between different time steps.
These kinds of predictions can lead to more conservative strategies within the subsequent motion planning~\cite{janson2018monte}.

\begin{figure*}
    \centering
    \resizebox{\textwidth}{!}{\input{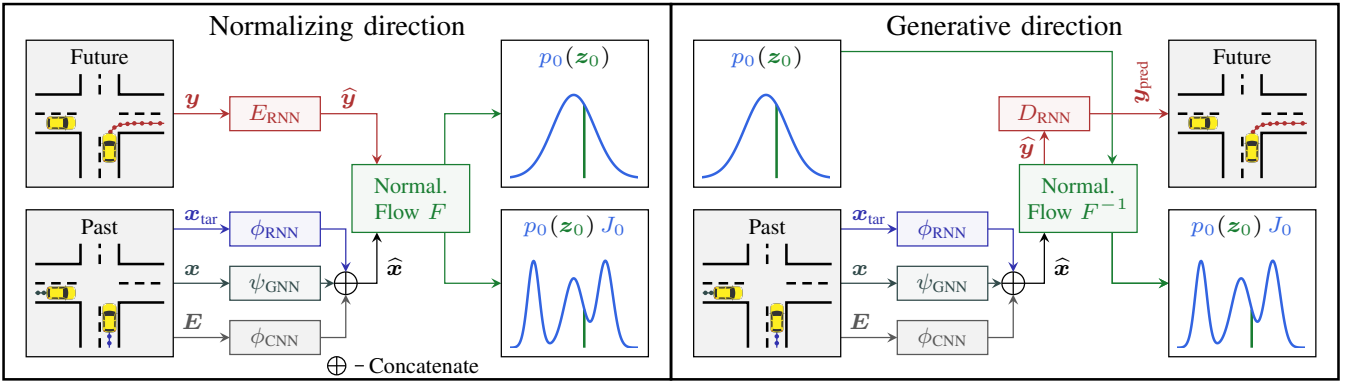}}
    \vspace{-4mm}
    \caption{Architecture of TrajFlow. 
    During training we use the normalizing direction in which we encode the future trajectories $\bm{y}$ with $E_{\text{RNN}}$ and transform the abstracted features $\widehat{\bm{y}}$ to a sample $\bm{z}_0 = F(\widehat{\bm{y}})$ assumed to follow a standard normal distribution with the probability density function $p_0$.
    For inference we then use the generative direction, in which a sample $\bm{z}_0 \sim p_0$ is inversely transformed by the Normalizing Flow to generate the abstracted future trajectories $\widehat{\bm{y}}= F^{-1}(\bm{z}_0)$ that are decoded with $D_{\text{RNN}}$ into the actual trajectories $\bm{y}_\text{pred}$.
    The likelihood of the encoded trajectory is obtained with $p_0 (\bm{z}_0) \vert \det J_{F^{-1}}(\bm{z}_0)\vert^{-1}$. 
    The encoding $\phi_{\text{CNN}}$ of map $E$, and the encoding $\psi_{\text{GNN}}$ of social interactions are optional blocks, which can provide richer context information. 
    }
    \vspace{-4mm}
    \label{fig:architecture}
\end{figure*}

To overcome these issues, one promising approach is Normalizing Flows (NFs)~\cite{tabak2010density, tabak2013family}, which are specifically designed to learn underlying distributions in data and have been shown to have the capability of capturing multi-modal distributions.
While NFs can be used to learn distributions of positions at individual time steps~\cite{rhinehart2018r2p2, rhinehart2019precog}, more recent work has expanded to providing distributions over complete trajectories~\cite{bhattacharyya2020haar, sun2021unified, scholler2021flomo}.
Even though the above NF methods already demonstrate good qualitative results in predicting multiple future trajectories, it remains unclear how well these models capture the true underlying distribution of the data.
Furthermore, the previously mentioned NF models which predict over the complete trajectories require one to set the number of predicted time steps during their design, which might limit their applicability and usefulness in an online setting.

In response to these challenges, the main contribution of this work is \emph{TrajFlow}---a prediction model with an improved capability for fitting distributions present in underlying training samples.
The model builds on top of \emph{FloMo}~\cite{scholler2021flomo}, which we extend with a key component in the form of a Recurrent Neural Network Autoencoder (Fig.~\ref{fig:architecture}).
This extension generates an intermediate representation of the trajectories, which captures the most relevant features of the trajectories and in turn also simplifies the learning of the underlying distribution.
The decoder of the Autoencoder is built in an auto-regressive manner, which additionally gives the model the flexibility to predict trajectories beyond the length of the seen training data. We validated our approach on a synthetic dataset for which we know the underlying distribution, as well as on an augmented version of the multi-modal Forking Paths dataset~\cite{liang2020garden}, and several popular real-world datasets (ETH/UCY~\cite{pellegrini2009you, lerner2007crowds}, rounD~\cite{rounDdataset}, and nuScenes~\cite{caesar2020nuscenes}).

\section{Background: Normalizing Flows}
Normalizing Flows constitute a family of generative methods which enable exact likelihood computation.
They are based on the concept of transforming distributions through a series of differentiable bijective functions into a simple known ``base'' distribution $Z_0$ -- most commonly a standard normal distribution.

A number of ways for constructing flow models have been proposed~\cite{papamakarios2021normalizing}.
One possible way is by using auto-regressive flows, consisting of a series of $K$ normalizing layers. 
The main components of these layers are the conditioner~$c_k$ and the transformer~$\tau_k$, which are often accompanied by an additional permutation layer~$\epsilon_k$.
The latter two functions ($\tau_k$ and $\epsilon_k$) are bijective -- and therefore invertible. 
In the generative direction, these functions then enable the transformation of a sample $\bm{z}_0$ from the base distribution $Z_0$ towards the desired distribution $Z_K$:
\begin{equation*}
    \bm{z}_{k+1} = \epsilon_k\left(\tau_k (\bm{z}_k; \bm{\theta}_k)\right), \quad \text{with} \quad \bm{\theta}_k=c_k\left(\bm{z}_{k};\bm{\widehat{x}}\right), \label{eq:NF_Layer}
\end{equation*}
where $\bm{z}_{k+1}$ is the result of the $k$-th intermediate transformation. 
Meanwhile, $\bm{\widehat{x}}$ is a conditioning input~\cite{lu2020structured} that can take the form of e.g. an encoding of observations like past trajectories, static environment, and social interactions.
In the normalizing direction, $F$ is then a composition of all $K$ layers, where it is possible to exploit the property of $c_k$ that $\bm{\theta}_k = c_k\left(\epsilon_k^{-1}(\bm{z_{k+1}});\bm{\widehat{x}}\right)$:
\begin{equation*}
    F(\bm{z}_K) = \left(\tau^{-1}_0 \circ \epsilon^{-1}_0 \cdots \circ \tau^{-1}_K \circ \epsilon^{-1}_K \right) (\bm{z}_K) = \bm{z}_0
\end{equation*}

In the generative direction, this then allows the drawing of a sample ${\bm{z}_K = F^{-1}(\bm{z}_0)}$ from the desired non-normal distribution over outputs $Z_K$, using a sample $\bm{z}_0$ from $Z_0$. 
The Probability Density Function (PDF) $p_K$ of $Z_K$ can then also be obtained in terms of the PDF $p_0$:
\begin{equation*}
\begin{split}
    p_K(\bm{z}_K) &= p_0(F(\bm{z}_K)) \, |\det J_{F}(\bm{z}_K)| \\
                    &= p_0(\bm{z}_0) \, |\det J_{F^{-1}}(\bm{z}_0)|^{-1},
\end{split}
\end{equation*}
The absolute determinant of the Jacobian $|\det J_F(\bm{z}_K)|$ quantifies the relative change of volume within a small neighborhood of $\bm{z}_K$ when transforming it to a sample $\bm{z}_0$ using $F$.
This ensures that the probability mass remains the same between the two distributions.

The parameters of $F$ are learned by minimizing the KL-divergence between the target distribution~$Z_K^{*}$ with PDF $p_K^{*}(\bm{z}_K)$ and the learned distribution $Z_K$ with the PDF $p_K(\bm{z}_K)$:
\begin{align*}
    \mathcal{L} =& D_{\mathrm{KL}}[p_K^{*}(\bm{z}_K)||p_K(\bm{z}_K)]\\
                =& -\mathbb{E}_{\bm{z}_K\sim Z_K^*}\left[\log p_0(F(\bm{z}_K)) +\log |\det J_{F}(\bm{z}_K)| \vphantom{\log p_K^{*}(\bm{z}_K)}\right. \\
                &\phantom{-\mathbb{E}_{\bm{z}_K\sim Z_K^*}}
                \; \left. \vphantom{\log p_0(F(\bm{z}_K)) +\log |\det J_{F}(\bm{z}_K)|}
                - \log p_K^{*}(\bm{z}_K)\right]\notag\\ 
\intertext{With only a finite number $N$ of samples $\bm{z}_{K, n}$ representing the underlying distribution $Z_K^{*}$ and ignoring the constant part $\log p_K^{*}(\bm{z}_K)$, this loss can be approximated with: }
    \mathcal{L} \approx& -\frac{1}{N}\sum^{N}_{n=1}\log p_0(F(\bm{z}_{K,n})) + \log |\det J_{F}(\bm{z}_{K,n})|.
\end{align*}

\section{Method: TrajFlow}
\label{sec3}
We build up on the \emph{FloMo} approach~\cite{scholler2021flomo}, in which NFs are employed for learning distributions directly on two-dimensional trajectories $\bm{y}~\in~\mathbb{R}^{n_O \times 2}$ defined at $n_O$ future time steps (where $\bm{y} = \bm{z}_K$). 
However, attempting to learn a distribution over the trajectories directly makes it susceptible to overfitting to the variability inherent in human behavior~\cite{siebinga2023uncovering} as well as noise in its measurements. 
Additionally, as $n_O$ in this design is fixed, the model has a limited prediction horizon, hindering its general applicability.
Furthermore, intuitively people do not observe trajectories as a series of precise positions at each time step.
Instead, they perceive a trajectory more abstractly in terms of general direction, length, and shape. 
Therefore, learning the distribution over such abstracted characteristics might be better suited to mimic human decision making, an approach that has shown itself to be promising in improving prediction models~\cite{cao2022leveraging, song2022research}.

To overcome these challenges and to facilitate the learning of underlying distributions, we constructed our proposed model \emph{TrajFlow} to let the NFs reason over trajectory abstractions rather than the raw trajectories. 

\subsection{Normalizing Flow}
\label{sec: normalizing flow TF}
In our specific case, we chose to use a \emph{Coupling Layer} for $c_k$, a \emph{Rational Quadratic Spline} for $\tau_k$, and a permutation layer~$\epsilon_k$, similar to the \textit{FloMo} model~\cite{scholler2021flomo}. 
However, unlike~\cite{scholler2021flomo}, we did not augment the trajectories in the training data. 
Furthermore, we also did not inject added noise into the Normalizing Flow as was done in FloMo using the $\beta$ and $\gamma$ hyperparameters.
Lastly, we incorporated a learning rate decay $lr$ for the training of the Normalizing Flow, as this has proven beneficial for achieving a better distribution fit.

\subsection{Encoding Trajectories}
To capture the abstracted characteristics of a trajectory, we utilized a Recurrent Neural Network Autoencoder (RNN-AE) with encoder $E_{\text{RNN}}$ and decoder $D_{\text{RNN}}$. 
This allows us to create an abstraction of a trajectory $\bm{\widehat{y}} = E_{\text{RNN}}(\bm{y}) \in \mathbb{R}^m$. 
This addition results in the novel \emph{TrajFlow} model (Fig.~\ref{fig:architecture}) that consequently learns the distribution of the encoded future trajectories $\widehat{Y}$ (with $\bm{\widehat{y}} = \bm{z}_K$) rather than the raw future trajectories $Y$.

\subsubsection{Gated Recurrent Unit}
The RNN-AE uses as its main component a so-called Gated Recurrent Unit (GRU)~\cite{cho-etal-2014-learning}, one of the main RNNs used for encoding time series events. 
In its most basic single-layered form with embedding dimensionality $M$ and hidden dimensionality $d$, it can be depicted as a function $\phi_{\text{GRU}}: \mathbb{R}^{M} \times \mathbb{R}^{d} \rightarrow \mathbb{R}^{d} $, which takes at time step $t$ an input $\bm{a}_t \in\mathbb{R}^M$ and uses it to change its internal hidden state $\bm{h}\in\mathbb{R}^d$:
\begin{equation*}
    \bm{h}_t = \phi_{\text{GRU}}\left(\bm{a}_t, \bm{h}_{t-1}\right)
\end{equation*} 
If no hidden layer is provided at the beginning of a sequence, those can be assumed to be zero. However, \emph{TrajFlow} employs a multi-layered version using multiple recurrent units $\phi_{\text{GRU}}^{(l)}$, with $l\in\{1,\hdots, L\}$:
\begin{equation*}
     \bm{h}_t^{(l)} = \phi_{\text{GRU}}^{(l)}   \left(\bm{h}_t^{(l - 1)}, \bm{h}_{t-1}^{(l)}\right) \quad \text{with} \quad \bm{h}_t^{(0)} = \bm{a}_t 
\end{equation*}
This can then be combined in a multilayer function $\phi_{\text{L-GRU}}: \mathbb{R}^{M} \times \mathbb{R}^{L\times d} \rightarrow \mathbb{R}^{L\times d}$ with $\bm{H}_t = \{\bm{h}_t^{(1)}, \hdots, \bm{h}_t^{(L)}\}$:
\begin{equation}
    \bm{H}_t = \phi_{\text{L-GRU}}\left(\bm{a}_t, \bm{H}_{t-1}\right) \label{eq:GRU}
\end{equation}

\subsubsection{The RNN-Encoder} \label{sec: RNN_encoder}
In the first step of the encoder $E_{\text{RNN}}$, we created a transformed trajectory $\bm{\widetilde{y}} = \langle\bm{\widetilde{y}}_1, \hdots, \bm{\widetilde{y}}_{n_O}\rangle$ with 
\begin{equation}
    \bm{\widetilde{y}}_t = \bm{y}_t - \bm{y}_{t-1} \label{eq:rel_pos}
\end{equation}
This is based on previous results showing that displacement information is more useful for trajectory prediction tasks~\cite{martinez2017human}.
We then used a linear layer $\phi_{\text{em}}: \mathbb{R} ^{2} \rightarrow \mathbb{R}^{M}$ that embeds a displacement $\bm{\widetilde{y}}_t$.
The embedded time steps are then run in sequence through a multi-layered GRU $\phi_{\text{E-L-GRU}}$~\eqref{eq:GRU}, setting $\bm{a}_t = \phi_{\text{em}}(\bm{\widetilde{y}}_t)$. Using a second linear layer $\phi_{\text{E}}: \mathbb{R} ^{d} \rightarrow \mathbb{R}^{m}$, we get our final encoded trajectory $\bm{\widehat{y}} = \phi_{\text{E}}\left(\bm{h}_{E,n_O}^{(L)}\right)$.

\subsubsection{The RNN-Decoder}
Our decoder $D_{\text{RNN}}$, uses as its first step a liner layer $\phi_{\text{D}}: \mathbb{R} ^{m} \rightarrow \mathbb{R}^{d}$ to pre-process an encoded trajectory $\bm{\widehat{y}}$. 
We then again used a multilayer GRU $\phi_{\text{D-L-GRU}}$ (Equation \eqref{eq:GRU}) to construct a new trajectory. Here, the initial hidden states are set to $\bm{h}^{(l)}_{D,0} = \bm{\widehat{y}}$. Meanwhile, our input is auto-regressive, i.e. $\bm{a}_1 = \phi_{\text{D}}(\bm{\widehat{y}})$ and $\bm{a}_t = \phi_{\text{D}}\left(\bm{h}_{D,t-1}^{(L)}\right)$ for $t > 1$. We constructed the final displacements using a linear layer $\phi_{\text{out}}: \mathbb{R} ^{d} \rightarrow \mathbb{R}^{2}$:
\begin{equation*}
    \bm{\widetilde{y}}_{t,\text{pred}} = \phi_{\text{out}}\left(\bm{h}^{(L)}_{D,t} \right)
\end{equation*}

As a last step, we used the cumulative sum over $\bm{\widetilde{y}}_{\text{pred}}$ to construct the predicted trajectory $\bm{y}_{\text{pred}}$ (inverting \eqref{eq:rel_pos}).
While we used the same hidden dimension $d$ and embedding size $M$ for both $\phi_{\text{E-L-GRU}}$ and $\phi_{\text{D-L-GRU}}$, we did not use any weight sharing between them. 

\subsubsection{Training}
The RNN-AE is trained separately before the rest of the network with a root mean square error reconstruction loss on the reconstructed trajectories:
\begin{equation*}
    \mathcal{L}_{\text{AE}} = {1\over{N}} \sum\limits_{n = 1}^N \Vert \bm{y}_n - \bm{y}_{\text{pred},n}\Vert.
\end{equation*}
The choice to calculate the loss on the reconstructed trajectories instead of the decoded displacements was made to penalize cumulative errors that can arise from summing over the displacements.
During the later training of the remaining parts of the model, the weights of the RNN-AE were frozen.  

\subsection{Encoding Context Information}

For the observations $\bm{\widehat{x}}$, which are used for conditioning the distributions learned by the NF, we used the target agent's past trajectory $\bm{x}_\text{tar}$, the past trajectories of all agents $\bm{x}$, and optionally images of the static environment $E$. In order to encode these pieces of information, we used the neural networks $\phi_{\text{RNN}}$, $\psi_{\text{GNN}}$, and $\phi_{\text{CNN}}$ respectively and concatenated their outputs:
\begin{equation*}
    \bm{\widehat{x}} = \phi_{\text{RNN}} (\bm{x}_\text{tar}) \oplus \psi_{\text{GNN}} (\bm{x}) \oplus \phi_{\text{CNN}} (\bm{E}) \label{eq:past_info}
\end{equation*}
The exact implementation of these components can be found in the Appendix.

\section{Experimental Setup}

We performed a number of tests, two on synthetic datasets with known ground truth distributions (Sec.~\ref{sec: synthetic examples}) and three on real-world datasets (Sec.~\ref{sec: real-world}).
To facilitate those tests, we utilized an existing benchmarking framework~\cite{schumann2023benchmarking}.

\subsection{Models}

We used four state-of-the-art behavior prediction models as baselines:
\begin{itemize}[align = left, labelsep = 0cm, labelwidth = \parindent, leftmargin = \parindent]
    \item \emph{Trajectron++ (T++)}~\cite{salzmann2020trajectron++}---selected as it continues to act as a strong baseline in trajectory prediction tasks. 
    At the same time, it provides a good illustration of the potential drawbacks of fitting distributions per time step.
    \item \emph{PECNet}~\cite{mangalam2020not}---a state-of-the-art model which captures multi-modality by predicting distributions over goal points and then regressing the trajectories to them.
    \item \emph{Motion Indeterminacy Diffusion (MID)}~\cite{gu2022stochastic}---a recent diffusion-based method for probabilistic trajectory prediction. 
    As of late, diffusion based models have been showing promise in generating probabilistic predictions~\cite{yang2023diffusion}.
    \item \emph{FloMo}~\cite{scholler2021flomo} (FM)---since we build on this model, it is most directly comparable to our approach.
    \item \emph{TrajFlow without the RNN-AE (TF w/o AE)}---to showcase the importance of the RNN-AE we tested against an ablation of TrajFlow.
\end{itemize}
For the RNN-AE in \emph{TrajFlow} we used a $L = 3$ layered GRU with a hidden dimensionality $d = 20$, embedding dimensionality $M = 20$, and latent space dimensionality $m = 20$ (Sec.~\ref{sec: RNN_encoder}).

For the past trajectory encoding $\phi_{\text{RNN}}$, we set the parameters in accordance to those described in the Appendix. 
Meanwhile, where applicable we employed the same $\psi_\text{GNN}$ and $\phi_\text{CNN}$ structures for \emph{TrajFlow}, \emph{TF w/o AE} and \emph{FloMo}.

The learning rate decay was set to $lr=0.98$.
These same parameters were used for \emph{TF w/o AE}.

\subsection{Metrics}
\label{sec:metrics}
To evaluate the distance of the predicted trajectories w.r.t.\ the ground truth trajectory we use:
\begin{itemize}[align = left, labelsep = 0cm, labelwidth = \parindent, leftmargin = \parindent]
    \item \textbf{minADE/minFDE}: Average/Final $L_2$ distance (measured in meters) between the best-predicted trajectory and the ground truth, based on $20$ predicted samples. We chose this metric primarily to obtain an interpretable measure of how closely the predictions of the models capture the single ground truth sample available in real-world test cases, since the usefulness of distribution specific metrics such as Negative Log-Likelihood (NLL) is limited when evaluating on singular ground truth samples.
\end{itemize}
In order to obtain insight into the learned distribution over trajectories, we make use of:
\begin{itemize}[align = left, labelsep = 0cm, labelwidth = \parindent, leftmargin = \parindent]
    \item $\mathbf{D_\text{JS}}$: Average Jensen-Shannon divergence~\cite{lin1991divergence} between the ground truth distribution and the learned distribution. 
    A perfect fit of the distribution is characterized by a divergence value of 0 whereas two dissimilar distributions would result in a divergence value of 1.
    As this metric requires a known ground truth distribution, it is only applicable for the synthetic cases.

    \item \textbf{(indep.) NLL}: The average NLL of the ground truth according to the learned distribution of each individual agent. 
    This gives us insight into the fit over the marginal distributions of the trajectories within the scene. This metric is commonly used in cases with any number of ground truth trajectories for a given scenario~\cite{salzmann2020trajectron++}.

    \item \textbf{joint NLL}: The average NLL
    of the ground truth, based on the joint distribution of predicted trajectories for all agents in the scene. This metric gives additional information about how well the model learned the interactions between the agents in the scene.
\end{itemize}
To obtain the density estimates needed for the above metrics, we used a non-parametric density estimation approach proposed in~\cite{mészáros2024robust} so as to ensure a more reliable comparison between models. For estimating the predicted trajectory distribution, $100$ sampled trajectories were used.

\section{Experiments: Synthetic Datasets}
\label{sec: synthetic examples}
\subsection{Datasets}
We tested our approach on two synthetically generated datasets, one of which was generated using a single bimodal distribution, while the other is an augmented version of the Forking Paths dataset~\cite{liang2020garden}.

\begin{figure*}
    \centering
    \resizebox{\textwidth}{!}{\input{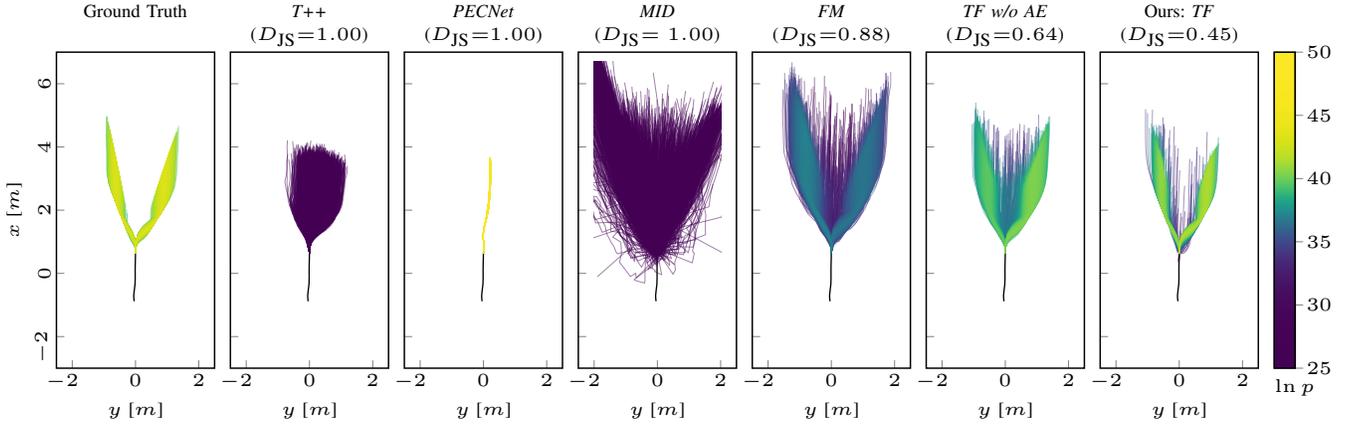}}
    \vspace{-2mm}
    \caption{Results of the experiments on the synthetic bimodal dataset. The left-most plot depicts the ground truth distribution; the other panels are the best out of the ten distributions learned by \emph{Trajectron++ (T++)}, \emph{PECNet}, \emph{Motion Indeterminacy Diffusion (MID)}, \emph{FloMo (FM)}, an ablation of \emph{TrajFlow} without the RNN-AE \emph{(TF w/o AE)}, and \emph{TrajFlow (TF)}, along with the $D_\text{JS}$ values for the specific distributions that are depicted.
    The colors provided in the distributions are determined based on the density values obtained through density estimation on 3000 samples obtained from the respective models. 
    }
    \vspace{-4mm}
    \label{fig:bimod_traj}
\end{figure*}
The \textbf{synthetic bimodal dataset} was used to test the models' ability to capture the underlying distribution of the observed data.
We constructed this dataset based on two recorded pedestrian trajectories with distinct directions to obtain a set of future trajectories over which we know the underlying distribution. This synthetic dataset has only a single agent and no static environment information for the observations.
The trajectories were split into a past sequence $\bm{x}^*_k$ with $n_I = 10$ and future sequence $\bm{y}^*_k$ with $n_O = 14$ recorded time steps of \SI{0.25}{s} each.
However, for both future trajectories, we used the same past trajectory $\bm{x}^*_1$ so that the true output distribution is guaranteed to be bimodal.
With this, we avoid the learned likelihoods becoming skewed due to slight differences in the past trajectories which could in turn make it more difficult to evaluate the predicted distributions. 
The set of future trajectories $\bm{Y}$ with $3000$ samples was then created by multiplying the original two future trajectories with a random scaling factor $s \sim \mathcal{N}(1, 0.15)$:
\begin{equation*}
    \bm{Y} = \left\{s_{1,i} \bm{y}^*_1, s_{2,i} \bm{y}^*_2 \; \vert \; i \in \{1,\hdots,1500\}      \right\} 
\end{equation*}


The \textbf{augmented Forking Paths dataset} was used to test the methods on a more complex, multi-modal dataset. 
The Forking Paths dataset~\cite{liang2020garden} originally included multiple human-predicted pedestrian trajectories. 
Within this dataset, for each past trajectory $\bm{x}^*$ with $n_I = 8$, there are $K$ annotated future trajectories $\bm{y}^*_k$ with $n_O = 12$, recorded with a sampling frequency of \SI{2.5}{Hz}.
We then generated $100$ augmented trajectories for each $k$:
\begin{equation*}
    \bm{y}_{k,i} = s_{k,i} \bm{y}^*_k \bm{R}_{\theta_{k,i}}^T \, , \, \text{with } k \in \{1,\hdots, K\}, i \in \{1, \hdots, 100\} \,. 
\end{equation*}
Here, $\bm{R}_\theta \in \mathbb{R}^{2\times 2}$ is a rotation matrix rotating $\bm{y}^*_k$ by $\theta \sim \mathcal{N}(0, \frac{\pi}{180})$, while $s \sim \mathcal{N}(1, 0.03)$ is a scaling factor.

\begin{table}
    \caption{Synthetic Bimodal Dataset: average results across 10 seeds}
    \centering
    \vspace{-1mm}

\setlength{\tabcolsep}{5pt}
\begin{tabular}{l|r @{\hskip 1pt} l|r@{\hskip 1pt}l|r@{\hskip 1pt}l|r@{\hskip 1pt}l}
\scriptsize  Models          & \multicolumn{2}{c|}{\scriptsize minADE} & \multicolumn{2}{c|}{\scriptsize minFDE} & \multicolumn{2}{c|}{\scriptsize NLL} & \multicolumn{2}{c}{\scriptsize  D$_\text{JS}$} \\ \hline
\scriptsize T++    & \scriptsize 0.43 &$\scriptscriptstyle{\pm 0.03}$ & \scriptsize 0.66 &$\scriptscriptstyle{\pm 0.09}$ & \scriptsize -19.08 &$\scriptscriptstyle{\pm 3.92}$ & \scriptsize 0.998 &$\scriptscriptstyle{\pm 0.001}$ \\
\scriptsize PECNet          & \scriptsize  0.90 &$\scriptscriptstyle{\pm 0.01}$ & \scriptsize  1.29 &$\scriptscriptstyle{\pm 4e^{-3}}$ & \scriptsize  $0.8e^3$ &$\scriptscriptstyle{\pm 10.77}$ & \scriptsize  1.000 &$\scriptscriptstyle{\pm 1e^{-16}}$\\
\scriptsize MID             & \scriptsize 0.72 &$\scriptscriptstyle{\pm 0.03}$ & \scriptsize 0.81 &$\scriptscriptstyle{\pm 0.20}$& \scriptsize \hphantom{0}-3.79 &$\scriptscriptstyle{\pm 1.88}$& \scriptsize 1.000 &$\scriptscriptstyle{\pm 5e^{-7}}$\\
\scriptsize FM           & \scriptsize 0.19 &$\scriptscriptstyle{\pm 0.03}$ & \scriptsize 0.32 &$\scriptscriptstyle{\pm 0.05}$& \scriptsize -35.83 &$\scriptscriptstyle{\pm 0.72}$& \scriptsize 0.916 &$\scriptscriptstyle{\pm 0.016}$ \\
\scriptsize TF w/o AE       & \scriptsize 0.13 &$\scriptscriptstyle{\pm 0.02}$& \scriptsize 0.22 &$\scriptscriptstyle{\pm 0.03}$ & \scriptsize -38.34 &$\scriptscriptstyle{\pm 1.05}$& \scriptsize 0.811 &$\scriptscriptstyle{\pm 0.081}$\\
\scriptsize TF (Ours) & \scriptsize \underline{0.12} &$\scriptscriptstyle{\pm 0.01}$ & \scriptsize \underline{0.20} &$\scriptscriptstyle{\pm 0.02}$& \scriptsize \underline{-39.22} &$\scriptscriptstyle{\pm 0.68}$& \scriptsize \underline{0.683} &$\scriptscriptstyle{\pm 0.132}$                 
\end{tabular} 

    \label{tab:Toy}
\end{table}

\subsection{Training and evaluation}
Considering that the \textbf{synthetic bimodal dataset} contains a single scenario, we trained $10$ instances of each model, using different random seeds, to decrease the effect of the random initialization of the models' trainable parameters.

Meanwhile, for the \textbf{augmented Forking Paths} dataset training and evaluation were performed using five-fold cross-validation, which was each repeated for $5$ random seeds.

\subsection{Results}
On the synthetic bimodal dataset, we found that out of the tested models, the methods which did not employ Normalizing Flows exhibited the poorest performance. This can be attributed to different factors, from the lack of correlation between time steps (\emph{T++}) to complete mode collapse (\emph{PECNet}).
The NF-based methods, meanwhile, were all able to capture the general shape of the underlying distribution.
Out of these, our approach \emph{TrajFlow (TF)} was able to provide the best fit.
A key factor to this is the use of the RNN-AE, which becomes clear when comparing the distributions learned by \emph{TF} with its ablation (Fig.~\ref{fig:bimod_traj}). 
These qualitative results are further supported by the low NLL and $D_\text{JS}$ values (Tab.~\ref{tab:Toy}) attained by \emph{TrajFlow}.

\begin{table}
    \caption{Forking Paths: Average results across all splits \& seeds.}
    \centering
    \vspace{-1mm}


\setlength{\tabcolsep}{5pt}
\begin{tabular}{l|r @{\hskip 1pt} l|r@{\hskip 1pt}l|r@{\hskip 1pt}l|r@{\hskip 1pt}l}
\scriptsize  Models          & \multicolumn{2}{c|}{\scriptsize minADE} & \multicolumn{2}{c|}{\scriptsize minFDE} & \multicolumn{2}{c|}{\scriptsize NLL} & \multicolumn{2}{c}{\scriptsize  D$_\text{JS}$} \\ \hline
\scriptsize T++    & \scriptsize 0.56 &$\scriptscriptstyle{\pm 0.06}$ & \scriptsize 1.02 &$\scriptscriptstyle{\pm 0.13}$ & \scriptsize -5.73 &$\scriptscriptstyle{\pm 4.21}$ & \scriptsize 0.985 &$\scriptscriptstyle{\pm 0.005}$ \\
\scriptsize PECNet          & \scriptsize 1.05 &$\scriptscriptstyle{\pm 0.09}$ & \scriptsize 1.89 &$\scriptscriptstyle{\pm 0.28}$ & \scriptsize $1.3e^3$ &$\scriptscriptstyle{\pm 0.5e^3}$ & \scriptsize 1.000 &$\scriptscriptstyle{\pm 1.3e^{-7}}$  \\
\scriptsize MID          & \scriptsize 0.89 &$\scriptscriptstyle{\pm 0.09}$ & \scriptsize 2.06 &$\scriptscriptstyle{\pm 0.38}$ & \scriptsize -7.41 &$\scriptscriptstyle{\pm 2.17}$ & \scriptsize 1.000 &$\scriptscriptstyle{\pm 5.7e^{-5}}$  \\
\scriptsize FM           & \scriptsize 0.41 &$\scriptscriptstyle{\pm 0.04}$ & \scriptsize 0.70 &$\scriptscriptstyle{\pm 0.10}$ & \scriptsize -21.86 &$\scriptscriptstyle{\pm 0.44}$ & \scriptsize 0.986 &$\scriptscriptstyle{\pm 0.003}$ \\
\scriptsize TF w/o AE   & \scriptsize \underline{0.40} &$\scriptscriptstyle{\pm 0.05}$ & \scriptsize \underline{0.69} &$\scriptscriptstyle{\pm 0.11}$ & \scriptsize -22.65 &$\scriptscriptstyle{\pm 0.69}$ & \scriptsize \underline{0.982} &$\scriptscriptstyle{\pm 0.004}$ \\
\scriptsize TF (Ours) & \scriptsize 0.42 &$\scriptscriptstyle{\pm 0.06}$ & \scriptsize 0.71 &$\scriptscriptstyle{\pm 0.11}$ & \scriptsize \underline{-23.69} &$\scriptscriptstyle{\pm 0.99}$ & \scriptsize 0.984 &$\scriptscriptstyle{\pm 0.004}$

\end{tabular} 

    \label{tab:ForkingPaths}
\end{table}

On the augmented Forking Paths dataset, we observed that none of the models could obtain distributions identical to the ones in the evaluation set -- as indicated by $D_\text{JS}$ values which are close to the maximum divergence value of $1$ (Tab.~\ref{tab:ForkingPaths}).
This is likely due to the fact that even though each of the past inputs has a ground truth distribution over the future trajectories, similar inputs may have different output distributions which could be merged by the models and thus result in dissimilar distributions from the actual ground truth distribution.
Nevertheless, NLL values show clear differences in the models' capability to capture the ground truth distributions; 
the Normalizing Flow methods are able to achieve better performance, with \emph{TrajFlow} achieving the best performance.

	

\section{Experiments: Real-world Datasets}
\label{sec: real-world}

\subsection{Datasets}
We tested our approach on three real-world datasets, ETH/UCY~\cite{pellegrini2009you, lerner2007crowds}, rounD~\cite{rounDdataset}, and nuScenes~\cite{caesar2020nuscenes}, all of which are widely used for trajectory prediction.

For testing the models on \textbf{ETH/UCY} (mostly including pedestrian crowds), we used $n_I = 8$ and $n_O = 12$ with a sampling frequency of \SI{2.5}{Hz}, resulting in \SI{3.2}{s} and \SI{4.8}{s} of past and future data respectively. 
Like in the majority of prior works, we did not make use of static environment information for the sake of comparability.

\begin{table}[t!]
    \caption{ETH/UCY: average results across the five locations.}
    \vspace{-1mm}
    \centering

\setlength{\tabcolsep}{5pt}
\begin{tabular}{l|r @{\hskip 1pt} l|r@{\hskip 1pt}l|r@{\hskip 1pt}l|r@{\hskip 1pt}l}
\scriptsize  Models          & \multicolumn{2}{c|}{\scriptsize minADE} & \multicolumn{2}{c|}{\scriptsize minFDE} & \multicolumn{2}{c|}{\scriptsize indep. NLL} & \multicolumn{2}{c}{\scriptsize  joint NLL} \\ \hline
\scriptsize T++  & \scriptsize 0.41&$\scriptscriptstyle{\pm 0.17}$ & \scriptsize 0.66&$\scriptscriptstyle{\pm 0.26}$ & \scriptsize -6.77&$\scriptscriptstyle{\pm 7.93}$ & \scriptsize $0.2e^3$
&$\scriptscriptstyle{\pm 0.5e^3}$ \\
\scriptsize PecNet        & \scriptsize 2.39&$\scriptscriptstyle{\pm 3.00}$ & \scriptsize 3.34&$\scriptscriptstyle{\pm 3.99}$ & \scriptsize $1.2e^4$ &$\scriptscriptstyle{\pm 2.2e^4}$ & \scriptsize $2.7e^4$ &$\scriptscriptstyle{\pm 4.3e^4}$\\
\scriptsize MID           & \scriptsize 0.59&$\scriptscriptstyle{\pm 0.16}$ & \scriptsize 1.08&$\scriptscriptstyle{\pm 0.30}$ & \scriptsize -0.27&$\scriptscriptstyle{\pm 10.21}$ & \scriptsize $0.1e^3$&$\scriptscriptstyle{\pm 0.2e^3}$ \\
\scriptsize FM         & \scriptsize \underline{0.32}&$\scriptscriptstyle{\pm 0.13}$ & \scriptsize 0.55&$\scriptscriptstyle{\pm 0.22}$  & \scriptsize -19.65&$\scriptscriptstyle{\pm 4.82}$ & \scriptsize -50.05&$\scriptscriptstyle{\pm 20.84}$ \\
\scriptsize TF w/o AE & \scriptsize 0.33&$\scriptscriptstyle{\pm 0.12}$ & \scriptsize \underline{0.54}&$\scriptscriptstyle{\pm 0.21}$ & \scriptsize -18.02&$\scriptscriptstyle{\pm 4.11}$ & \scriptsize -42.24&$\scriptscriptstyle{\pm 16.50}$ \\
\scriptsize TF (Ours) & \scriptsize 0.33&$\scriptscriptstyle{\pm 0.15}$ & \scriptsize 0.55&$\scriptscriptstyle{\pm 0.24}$ &  \scriptsize \underline{-21.05}&$\scriptscriptstyle{\pm 5.35}$ & \scriptsize \underline{-75.79}&$\scriptscriptstyle{\pm 47.42}$
                             
\end{tabular} 

    \label{tab:ETH}
\end{table}
For testing the models on \textbf{rounD} (drone-captured roundabouts), we set $n_I = 15$ and $n_O = 25$ with a sampling frequency of \SI{5}{Hz}, which amounts to \SI{3}{s} and \SI{5}{s} of past and future data respectively.
For our evaluation, we used the scenarios extracted from the original dataset as done in~\cite{schumann2023using}, which focused on the gap acceptance scenario of a vehicle entering the roundabout. 
There, both the trajectories of the vehicle entering the roundabout and the trajectory of the vehicle already inside the roundabout, which might be cut off by the former vehicle, have to be predicted. As it is important to predict if the other vehicle might yield when trying to plan for such scenarios, it can be necessary to predict more than $n_O = 25$ time steps. 
This is the case in 5.9\% of the scenes in rounD, with the longest predictions requiring 35 time steps.

Lastly, on \textbf{nuScenes} (general street traffic), we used $n_I=4$ and $n_O = 12$ with a sampling frequency of \SI{2}{Hz}. For both nuScenes and rounD, full context information is available.

\begin{table}[t!]
    \caption{RounD: average results across the five cross splits.}
    \vspace{-1mm}
    \centering

\setlength{\tabcolsep}{5pt}
\begin{tabular}{l|r @{\hskip 1pt} l|r@{\hskip 1pt}l|r@{\hskip 1pt}l|r@{\hskip 1pt}l}
\scriptsize  Models          & \multicolumn{2}{c|}{\scriptsize minADE} & \multicolumn{2}{c|}{\scriptsize minFDE} & \multicolumn{2}{c|}{\scriptsize indep. NLL} & \multicolumn{2}{c}{\scriptsize  joint NLL} \\ \hline
\scriptsize T++  & \scriptsize 0.69 &$\scriptscriptstyle{\pm 0.02}$ & \scriptsize 1.66 &$\scriptscriptstyle{\pm 0.07}$ & \scriptsize -42.04 &$\scriptscriptstyle{\pm 1.10}$ & \scriptsize -79.67 &$\scriptscriptstyle{\pm 1.98}$\\
\scriptsize PECNet        & \scriptsize 1.56 &$\scriptscriptstyle{\pm 0.11}$ & \scriptsize 4.49 &$\scriptscriptstyle{\pm 0.20}$ & \scriptsize $3.4e^3$ &$\scriptscriptstyle{\pm 0.3e^3}$ & \scriptsize $6.1e^3$ &$\scriptscriptstyle{\pm 0.7e^3}$ \\
\scriptsize MID           & \scriptsize 4.57 &$\scriptscriptstyle{\pm 0.01}$ & \scriptsize 7.93 &$\scriptscriptstyle{\pm 0.07}$ & \scriptsize $0.3e^3$ &$\scriptscriptstyle{\pm 0.1e^3}$ & \scriptsize $0.5e^3$ &$\scriptscriptstyle{\pm 0.2e^3}$ \\
\scriptsize FM         & \scriptsize 0.85 &$\scriptscriptstyle{\pm 0.03}$ & \scriptsize 1.80 &$\scriptscriptstyle{\pm 0.06}$ & \scriptsize -34.98 &$\scriptscriptstyle{\pm 3.65}$ & \scriptsize -69.35 &$\scriptscriptstyle{\pm 4.98}$ \\
\scriptsize TF w/o AE & \scriptsize 0.80 &$\scriptscriptstyle{\pm 0.01}$ & \scriptsize 1.72 &$\scriptscriptstyle{\pm 0.03}$& \scriptsize -36.86 &$\scriptscriptstyle{\pm 1.32}$ & \scriptsize -71.62 &$\scriptscriptstyle{\pm 2.43}$ \\
\scriptsize TF (Ours) & \scriptsize \underline{0.67} &$\scriptscriptstyle{\pm 0.03}$ & \scriptsize \underline{1.38} &$\scriptscriptstyle{\pm 0.09}$ & \scriptsize \underline{-42.06} &$\scriptscriptstyle{\pm 2.67}$ & \scriptsize \underline{-81.23} &$\scriptscriptstyle{\pm 4.69}$

\end{tabular} 

    \label{tab:RounD}
\end{table}

\subsection{Training and Evaluation}
For \textbf{ETH/UCY}, training and evaluation were performed using a leave-one-out strategy~\cite{salzmann2020trajectron++}, using the five recording locations (ETH-univ, ETH-hotel, UCY-univ, UCY-zara01, UCY-zara02).

For \textbf{rounD}, training and evaluation were performed using five-fold cross-validation.
While we evaluated the normal minADE metric based on the first 25 time steps, we also checked the models' capability to predict beyond that to test its ability of extending predictions until the point by which a yielding decision had to be reached.
If a model was unable to predict beyond the $25$ future time steps that it had been trained on, the values for the remaining time steps were obtained through a simple constant velocity extrapolation.

Lastly, training and evaluation for \textbf{nuScenes} were performed using nuScenes' predefined training and validation splits.
To decrease the effect of random parameter initialization, we trained $5$ different versions of each model, using different random seeds.

\subsection{Results}
On ETH/UCY, we observed the same trend as in the synthetic datasets.
The three NF-based methods were able to better fit the underlying data distribution compared to the methods which do not employ NFs.
Out of these, \emph{TrajFlow} clearly outperformed existing methods as well as its ablation in terms of distribution fit as captured by both NLL metrics. (Tab.~\ref{tab:ETH}) 
This is especially clear in the joint NLL, indicating that the learned distributions were also better able to capture the interactions among agents in a scene.
We found that the state-of-the-art methods also performed worse even compared to the originally reported values, such as in the case of \emph{T++}~\cite{salzmann2020trajectron++}.
It is, however, important to note that this is not the first time the original \emph{T++} results could not be replicated (see e.g.~\cite{westerhout2023smooth}), and compared to common practice, we used a stricter method for extracting testing samples, necessitating the existence of all $12$ future positions.

On rounD (Tab.~\ref{tab:RounD}), \emph{TrajFlow} was able to achieve the best results. 
When compared to its ablation case, there is a clear performance boost both in terms of the learned distribution but also in terms of the distance metrics. 
This is especially notable since in rounD the agents being predicted are vehicles, not pedestrians.
As a result, the distance crossed is generally larger and thus prediction errors tend to accumulate faster. 
In terms of the extrapolation performance, out of all the tested models, \emph{TrajFlow} achieved the best minADE value with \SI{2.27}{m}$^{\pm0.30}$. 
This was closely followed by \emph{TF w/o AE} and \emph{T++} with values of \SI{2.53}{m}$^{\pm0.52}$ and \SI{2.58}{m}$^{\pm0.55}$ respectively.
\emph{FloMo} achieved an error of \SI{2.62}{m}$^{\pm0.59}$, while \emph{MID} and \emph{PECNet} achieved an error of \SI{10.10}{m}$^{\pm1.26}$ and \SI{19.09}{m}$^{\pm1.10}$ respectively.
It is worth noting that out of these methods, only \emph{TrajFlow} and \emph{T++} have auto-regressive capability while the remaining models require a separate extrapolation method.

\begin{table}[t!]
    \caption{NuScenes: validation split results across the five seeds.}
    \vspace{-1mm}
    \centering

\setlength{\tabcolsep}{5pt}
\begin{tabular}{l|r @{\hskip 1pt} l|r@{\hskip 1pt}l|r@{\hskip 1pt}l|r@{\hskip 1pt}l}
\scriptsize  Models          & \multicolumn{2}{c|}{\scriptsize minADE} & \multicolumn{2}{c|}{\scriptsize minFDE} & \multicolumn{2}{c|}{\scriptsize indep. NLL} & \multicolumn{2}{c}{\scriptsize  joint NLL} \\ \hline
\scriptsize T++  & \scriptsize 1.96 &$\scriptscriptstyle{\pm 0.02}$ & \scriptsize 4.05 &$\scriptscriptstyle{\pm 0.07}$ & \scriptsize \underline{10.28} &$\scriptscriptstyle{\pm 0.29}$ & \scriptsize \underline{43.18} &$\scriptscriptstyle{\pm 1.01}$ \\
\scriptsize PECNet        & \scriptsize 26.73 &$\scriptscriptstyle{\pm 2.30}$ & \scriptsize 40.04 &$\scriptscriptstyle{\pm 2.36}$ & \scriptsize $6.7e^5$ &$\scriptscriptstyle{\pm 1.0e^5}$ & \scriptsize $1.8e^6$ &$\scriptscriptstyle{\pm 2.9e^5}$ \\
\scriptsize MID           & \scriptsize 6.47 &$\scriptscriptstyle{\pm 0.29}$ & \scriptsize 13.48 &$\scriptscriptstyle{\pm 0.82}$ & \scriptsize 71.04 &$\scriptscriptstyle{\pm 7.97}$ & \scriptsize $0.3e^3$ &$\scriptscriptstyle{\pm 23.27}$ \\
\scriptsize FM         & \scriptsize 12.61 &$\scriptscriptstyle{\pm 1.07}$ & \scriptsize 23.55 &$\scriptscriptstyle{\pm 2.02}$ & \scriptsize $0.3e^3$ &$\scriptscriptstyle{\pm 0.2e^3}$ & \scriptsize $0.9e^3$ &$\scriptscriptstyle{\pm 0.7e^3}$ \\
\scriptsize TF w/o AE & \scriptsize 13.15 &$\scriptscriptstyle{\pm 0.18}$ & \scriptsize 24.71 &$\scriptscriptstyle{\pm 0.29}$ & \scriptsize $0.9e^3$ &$\scriptscriptstyle{\pm 0.6e^3}$ & \scriptsize $1.9e^3$ &$\scriptscriptstyle{\pm 1.3e^3}$ \\
\scriptsize TF (Ours) & \scriptsize \underline{1.86} &$\scriptscriptstyle{\pm 0.06}$ & \scriptsize \underline{3.72} &$\scriptscriptstyle{\pm 0.19}$ & \scriptsize 16.54 &$\scriptscriptstyle{\pm 1.04}$ & \scriptsize 45.55 &$\scriptscriptstyle{\pm 2.72}$ 
                             
\end{tabular} 

    \label{tab:nuScenes}
\end{table}

Finally, on nuScenes (Tab.~\ref{tab:nuScenes}), \emph{TrajFlow} outperformed all of the models in terms of the distance metrics minADE and minFDE.
It was, however, outperformed on the distribution metrics by \emph{T++}.
What is notable, however, is that although \emph{TrajFlow} had poorer performance than \emph{T++} on the independent NLL, the performance of the two models on the joint NLL was comparable.
This indicates that while \emph{TrajFlow} does not capture the individual distributions as precisely as \emph{T++} on nuScenes, the distributions learned manage to reflect the overall behavior in a scene to a similar extent as \emph{T++}.

\section{Conclusion}
\label{sec:conclusion}
\label{sec6}
In this work, we proposed \emph{TrajFlow}, a novel model for predicting the trajectories of human agents in traffic by applying Normalizing Flows to the latent 
abstraction of the future trajectories to be predicted.

Tests carried out on synthetic examples, which contained sets of grounds truths for a single input, showed that Normalizing Flow-based methods outperformed several state-of-the-art methods in terms of the distribution fits.
Furthermore, among these NF-based models, \textit{TrajFlow} had the best performance thanks to incorporating a Recurrent Neural Network-based Autoencoder.

Through evaluations on the ETH/UCY, rounD, and nuScenes datasets, we found that our model can successfully learn distributions within real-world datasets.
\emph{TrajFlow} achieved competitive results compared to state-of-the-art methods on all three datasets, and was able to clearly outperform existing methods in terms of the distribution fit on the pedestrian dataset ETH/UCY.
This is particularly noteworthy since pedestrian behavior is less structured than that of vehicles and one can in turn expect a higher amount of variability. 

We further observed that the introduction of an RNN-AE does not lead to a severe degradation of our predicted trajectories despite the loss of information inherent to data compression, and in fact, for rounD and nuScenes, learning over abstracted trajectory features proved to be beneficial.
Tests on rounD also showed that the auto-regressive nature of our decoder provides further flexibility in terms of the possible length of the predictions and is even able to outperform state-of-the-art models in terms of prediction accuracy.
This is particularly useful for cases that need a longer planning horizon to ensure safety and comfort such as when approaching a roundabout or when interacting with pedestrians close to a crosswalk.

Future work will explore ways to improve prediction quality for more structured environments such as in the case of vehicle trajectory prediction. 
Another point of focus will be to expand the model towards being able to provide joint predictions of all agents in a given scene. Finally, \textit{TrajFlow}'s capacity to capture multi-modal distributions can be utilized in contingency planners which account for multiple possible outcomes \cite{rhinehart2021contingencies}. The extent to which better distribution fitting is beneficial to such planners was outside of the scope of this work and should be investigated in future work.

Overall, our results indicate that \emph{TrajFlow} compares favorably to state-of-the-art behavior prediction models in learning trajectory distributions both from highly variable data (e.g., pedestrian trajectories) as well as more constrained scenarios (such as in the case of vehicle trajectories). This model thus has potential for application in settings where AVs have to navigate in settings with human road users in order to generate more natural and safer plans.



\section*{APPENDIX}

\section*{The Encoding of Past Behavior}
\label{app:enc_modules}
\subsection{Past Trajectories}
\label{app:enc_modules_past}
The encoder of the past behavior $\phi_{\text{RNN}}$ encodes the past trajectory $x_\text{tar}$ of the single target agent whose future is to be predicted. This function was taken from the implementation of~\cite{scholler2021flomo} and is identical to the encoder $E_{\text{RNN}}$ (see Sec.~\ref{sec: RNN_encoder}), except that instead of $d = M = m = 20$, we used $d = M = m = 64$ for the TrajFlow variants and $d = M = m = 16$ in FloMo, as per the original implementation.

\subsection{Static Environment}
For encoding a gray-scale image of the static environment $E$, which has been rotated to align with the target agent's heading, we used a CNN function $\phi_{\text{CNN}}$.
For this, $L_{\text{CNN}} = 3$ convolutional layers $\phi_{\text{CNN}}^{(l)}$ are used within this network with a kernel of size $5$ and a stride of $4$. 
The first two layers additionally have a zero-padding of size $1$ around the image. 
With this, an initial input of size $h^{(0)} \times w^{(0)} = 156 \times 257$ and $c^{(0)} = 1$ channel is transformed first into a representation with $c^{(1)} = 8$ and $h^{(1)} \times w^{(1)} = 39 \times 64$, then into a representation with $c^{(2)} = 16$ and $h^{(2)} \times w^{(2)} = 10 \times 16$ and lastly into an output representation with $c^{(3)} = 32$ and $h^{(3)} \times w^{(3)} = 2 \times 3$. 
This output $\phi_{\text{CNN}}^{(3)}$ is then flattened and passed through a two-layer dense network. 
The first linear layer transforms the input into a hidden state of length $128$, while the second linear layer produces the final encoding of the image of size $M_{\text{CNN}}=64$.

\subsection{Social Interactions}
To encode interactions, we use a GNN function $\psi_{\text{GNN}}$ that processes all past trajectories $\bm{x} = \left\{\bm{x}_\text{tar}, \bm{x}_1, \hdots \bm{x}_{n-1}\right\}$.
There, in the first step, the past trajectory $\bm{x}_a$ of each agent $a$ of the $n$ agents is encoded using a GRU-based function $\psi_{\text{RNN}, c}$ (structure is identical to $\phi_{\text{RNN}}$; Sec.~\ref{app:enc_modules_past}). 
This network is shared between all agents of each class $c \in C = \{\text{veh.}, \text{ped.}, \hdots\}$, i.e. there is for instance one network $\psi_{\text{RNN, veh.}}$ to encode the past of vehicles. 
A linear embedding layer $\psi_{\text{em}}: \mathbb{R}^m \rightarrow \mathbb{R} ^ {M_{\text{GNN}}}$ is then applied to each encoded past trajectory, with $\widetilde{\bm{x}}_a^{(0)} = \psi_{\text{em}}\left(\psi_{\text{RNN},c_a}\left(\bm{x}_a\right)\right)$. 

In the GNN, each of the $n$ agents is seen as a node, with $n^2$ unidirectional edges being established between all nodes. Based on this, $L_{\text{GNN}}$ layers $\psi^{(l)}_{\text{GNN}}$ are applied to this network to update the node states $\widetilde{\bm{x}}^{(l)} = \left\{\widetilde{\bm{x}}^{(l)}_\text{tar}, \widetilde{\bm{x}}^{(l)}_1, \hdots, \widetilde{\bm{x}}^{(l)}_{n-1}\right\}$: 
\begin{equation*}
    \widetilde{\bm{x}}^{(l)} = \psi^{(l)}_{\text{GNN}} \left(\widetilde{\bm{x}}^{(l-1)} \right)
\end{equation*}
The update starts with calculating the message $\bm{m}^{(l)}_{b,a}$ from agent $b$ to agent $a$ for every possible connection, using the message network $\psi_{\text{M}}:\mathbb{R} ^{2\, M_{\text{GNN}} + 2\, \vert C\vert + 1} \rightarrow \mathbb{R} ^ {M_{\text{GNN}}}$:
\begin{equation*}
    \bm{m}^{(l)}_{b,a} = \psi_{\text{M}}\left( 
    \widetilde{\bm{x}}^{(l - 1)}_b \oplus \\\widetilde{\bm{x}}^{(l - 1)}_a \oplus \mathcal{C}_b \oplus \mathcal{C}_a \oplus \left\Vert \bm{x}_a - \bm{x}_b \right\Vert\right),
\end{equation*}
where the last three terms are the graph's edge features between agents $a$ and $b$, with $\mathcal{C}_a, \mathcal{C}_b \in \mathbb{R}^{\vert C\vert}$ being the one-hot encoding of class $c_a$ and $c_b$ respectively. 
Those incoming messages are then aggregated at each node:
\begin{equation*}
    \bm{m}_a^{(l)} = \sum\limits_b \bm{m}^{(l)}_{b,a}
\end{equation*}
Lastly, the state of each node is updated, using an update network $\psi_{\text{U}}:\mathbb{R} ^{2\, M_{\text{GNN}}} \rightarrow \mathbb{R} ^ {M_{\text{GNN}}}$
\begin{equation*}
    \widetilde{\bm{x}}^{(l)}_a = \psi^{(l)}_{\text{U}} \left(\widetilde{\bm{x}}_a^{(l-1)} \oplus \bm{m}_a^{(l)} \right) + \widetilde{\bm{x}}_a^{(l - 1)}
\end{equation*}

After being propagated through all $L_{\text{GNN}}$ layers $\psi^{(l)}_{\text{GNN}}$, the final output of $\psi_{\text{GNN}}$ is
\begin{equation*}
    {1\over n} \sum\limits_i  \widetilde{\bm{x}}^{(L_{\text{GNN}})}_a
\end{equation*}

For our work, we chose to set $L_{\text{GNN}} = 4$ and $M_{\text{GNN}} = 32$.

\bibliographystyle{jabbrv_ieeetr}
\bibliography{IEEEabrv,IEEEexample}  

\end{document}